%% file: main.tex
\documentclass{article} 
\usepackage{latticelab,times}

\nonanonymous

\input{math_commands.tex}

\usepackage{hyperref}
\usepackage{url}
\usepackage{graphicx}
\usepackage{amssymb}
\usepackage{booktabs} 
\usepackage{algorithm}
\usepackage{algpseudocode}
\usepackage{tcolorbox}
\usepackage{multirow}
\usepackage{booktabs}
\usepackage{amsmath}

\RequirePackage{graphicx}
\RequirePackage{eso-pic}
\usepackage{xcolor}         
\usepackage{todonotes}   
\usepackage{paralist} 

\title{LapidaryEngine: Fully Conversational Crystal Generation}
\shorttitle{LapidaryEngine: Fully Conversational Crystal Generation}

\paperlogo{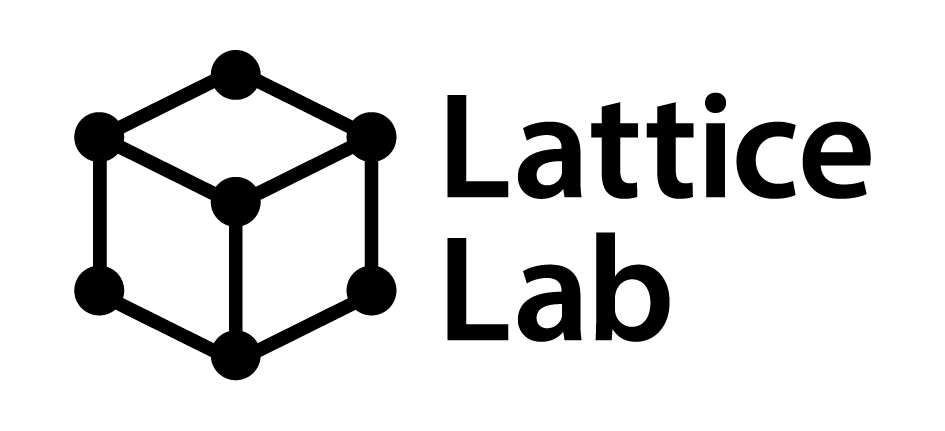}

\author{
Yusei Ito$^{1,2}$\thanks{The work was done while the first author was an intern at Lattice Lab, Toyota Motor Corporation.}
\quad
Yuta Suzuki$^{1}$
\quad
Tomoya Murata$^{1}$\thanks{Corresponding author: \href{mailto:tomoya_murata_aa@mail.toyota.co.jp}{\nolinkurl{tomoya_murata_aa@mail.toyota.co.jp}}}
\quad
Masaki Adachi$^{1}$
\\[0.5em]
{$^{1}$Lattice Lab, Toyota Motor Corporation, $^{2}$The University of Osaka}
}

\newcommand{\ie}{\emph{i.e.}}
\newcommand{\eg}{\emph{e.g.}}

\newcommand{\methodname}{LapidaryEngine}

\begin{document}

\maketitle

\begin{abstract}
The emergence of Large Language Models (LLMs) has inspired the vision of generating bespoke crystal materials directly from natural-language instructions, enabling users to design materials through intuitive, conversational interaction. Existing text-to-crystal generative models represent important early steps toward this goal, but they suffer from two critical limitations: (i) restricted input formats that require highly structured descriptions (e.g., chemical formulas), and (ii) one-directional generation, where models can map text $\to$ crystal but cannot perform the inverse. These limitations prevent fully conversational workflows and hinder alignment with users’ inherently ambiguous and evolving desiderata.
We address these challenges with \methodname, the first model to support fully conversational crystal generation. \methodname~accepts free-form natural-language requests and performs iterative refinement and editing in a dialogue-like manner. The key innovation is a pivot representation—a third, intermediate form that enables bidirectional translation between text and crystal structures despite the absence of direct paired datasets. Leveraging this pivot allows robust interpretation of user feedback and precise structural control.
We demonstrate \methodname~across diverse tasks, including insulator discovery, stability optimization, compositional modification, and structural editing, showcasing its ability to align generated materials with user intent in an interactive manner.
\end{abstract}

\section{Introduction} \label{s:introduction}
Given the remarkable success of generative models in image~\citep{rombach2021stablediffusion}, video~\citep{2024SoraReview}, and music synthesis~\citep{agostinelli2023musiclm}, it is natural to expect the recent breakthroughs in generative modeling to extend to materials design—indeed, the number of AI-for-materials papers has rocketed dramatically~\citep{Wang2023ai-for-science-review, Stokes2020halcin, Jumper2021alphafold, Merchant2023gnome, Zeni2025MatterGen}. In particular, with the emergence of human-level performance in Large Language Models (LLMs)~\citep{OpenAI2023GPT4, gemini2023gemini, Yang2025qwen3}, the community is now racing to apply these capabilities to core challenges in scientific fields, including hypothesis generation, experimental planning, and automated scientific reasoning~\cite{lewkowycz2022solving, chris2024aiscientist, abbi2025scientific, Swanson2025virtuallab}.
One of the most transformative aspects of LLMs is the ability to formulate scientific problems directly in natural language—tasks that previously required carefully crafted, domain-specific formalisms for simulations and experiments. Among these new possibilities, text-to-crystal generation stands out: it offers an interface through which users can specify desired materials simply by writing natural-language descriptions. Because traditional materials-science tools—such as atomistic simulations or quantum-chemical analysis—have long been inaccessible to non-experts, text-to-crystal systems hold the promise of democratizing expert knowledge. Much like how non-experts can now write novels or create illustrations with generative models, engineers and designers could soon generate bespoke materials tailored to their needs.

Two early works, Chemeleon and GenMS, represent pioneering steps toward text-to-crystal models, demonstrating promising results in their respective applications~\citep{yang2024genms, Park2025Chemeleon}. Chemeleon shows that crystals can be generated from only a vague textual prompt describing the composition, and in the Zn-Ti-O system it explores the chemical space to propose new candidate crystal structures beyond the known phases. GenMS demonstrates that natural language descriptions specifying a crystal family, such as perovskites, can be used to correctly generate crystal structures consistent with that family.
However, these approaches do not yet achieve the goal of democratization. Why? Because crafting precise instructions itself requires expert-level or even oracle-level knowledge. Natural-language descriptions can be inherently ambiguous. For example, consider the prompt: “Generate an insulating material that has not yet been reported in the literature.” The space of possible answers is vast, leaving open questions about the material family, the required degree of insulation (e.g., bandgap), and various physical or chemical constraints. To specify these details, users must already possess a clear and technically grounded design target.
Moreover, at the scientific frontier, even experts operate under uncertainty: they often do not know whether a hypothesized structure can be synthesized or whether it will exhibit the desired properties. Their design goals typically evolve through iterative trial and error. Requiring users to provide well-specified, error-free instructions a priori therefore conflicts with how real materials discovery proceeds and severely limits applicability.

A more natural and user-friendly interface for materials design is iterative refinement, exemplified by the conversational paradigm popularized by ChatGPT. Users should be able to begin with a vague idea, progressively clarify their requirements as they learn more, and ultimately converge to a well-specified target through a dialogue-like interaction.
However, enabling \emph{conversational} crystal generation is far from trivial. The two existing text-to-crystal systems \citep{yang2024genms, Park2025Chemeleon} cannot support such interactive refinement. This stems from two fundamental limitations:
\begin{compactenum}[(a)]
    \item \textbf{Restricted instruction formats.}
    Existing approaches typically constrain textual inputs to highly structured descriptions—such as chemical formulas, space groups, or symmetry tokens~\citep{Park2025Chemeleon}. These rigid formats deviate substantially from natural language, requiring expert-level knowledge to construct valid prompts. Moreover, they exclude intentionally ambiguous or partially specified requests, which are essential for true democratization and early-stage ideation.
    \item \textbf{One-directional modeling.}  
    Current methods map \emph{text} $\to$ \emph{crystal}, but not \emph{crystal} $\to$ \emph{text}~\citep{yang2024genms, Park2025Chemeleon}. Without a reverse pathway, the model cannot interpret the previously generated structure, assess how it aligns with the user’s evolving intent, or incorporate feedback from earlier rounds. As a result, iterative refinement is impossible: the system has no mechanism to update or adjust the design based on the output of prior iterations.
\end{compactenum}

\begin{figure}[t!]
\centering
\includegraphics[width=1.0\textwidth]{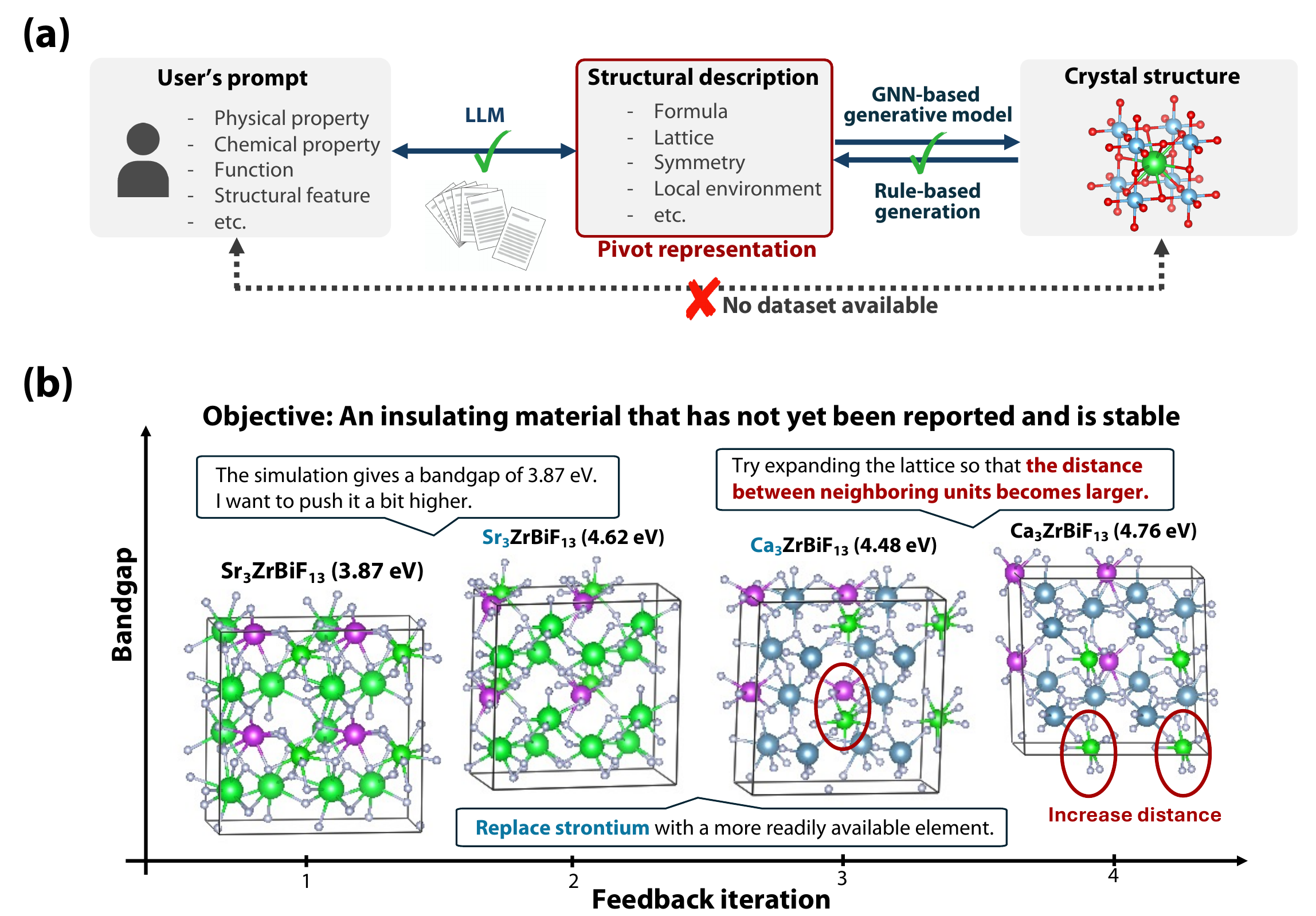}
\caption{
\textbf{Key idea and example of \methodname}. (a) As there is no dataset that directly links textual descriptions to crystal structures, our approach introduces a pivot representation bridging linguistic and structural modalities.
(b) Examples of crystal structures generated by our framework. Starting from a prompt requesting an insulator with a large bandgap, the model produces an initial structure and iteratively refines it through natural language feedback.
}
\label{fig:teaser}
\end{figure}

In response, we propose \methodname—the \emph{first-of-its-kind} model to enable \emph{fully} conversational, multi-round refinement of crystal structures. Our key innovation is the introduction of \emph{a pivot representation}, inspired by classical pivot-based machine translation. Just as two languages without parallel data (e.g., Kiswahili and Japanese) can communicate via a shared third language (English), we establish a pivot representation that bridges text and crystal structure.
Crucially, while direct datasets of (property, crystal) pairs do not exist, both modalities can be bidirectionally mapped to a structural description—our pivot representation. As illustrated in Fig.~\ref{fig:teaser}(a), this pivot provides a common semantic ground through which text and crystal structures become mutually interpretable, enabling stable, iterative, conversational refinement for the first time.

We demonstrated \methodname~on diverse tasks. Figure~\ref{fig:teaser}(b) shows the main result. Starting from a prompt that requests an insulating material (i.e., one with a large bandgap), an initial structure is generated as a rough hypothesis. The system then refines this structure through iterative user feedback, beginning with coarse guidance and progressively incorporating the constraints that naturally arise during crystal structure design. This framework is not posed as a conventional optimization task. Instead, it allows the user’s desiderata, including preferences, design intentions, and domain-specific considerations, to be continuously injected and reflected in the evolving structure. In addition, we conducted two tasks aimed at improving verifiable physical properties. Each task was repeated 1,000 times, and statistical analysis confirmed that the targeted physical properties were improved.
We open-source our code and model to the community.

\section{Results} \label{s:results}
Toward fully conversational crystal generation, we show that a pivot representation based on structural descriptions resolves both flexibility and bidirectionality challenges. The key idea is simple: instead of directly editing the crystal structure—which is discrete, highly constrained, and difficult to manipulate—we edit a textual description that represents the structural information. This allows all refinement steps to remain entirely within the text domain, where LLMs excel at controlled editing and iterative improvement. By shifting the problem from the joint (text, crystal) space to the text-only pivot space, we can fully exploit the strengths of LLMs while maintaining precise control over the generated structures. We demonstrate that this design solves the limitations that existing methods fail to overcome.

\subsection{Pivot representation} \label{s:pivot_representation}
Although no generative model currently supports \emph{crystal}~$\to$~\emph{text}, there exists a \emph{rule-based} text generator for crystal structures: Robocrystallographer \citep{Ganose2019Robocrystallographer}. Crucially, because it is rule-based, its mapping is essentially bijective: a crystal has one corresponding textual description, and vice versa. This property makes the Robocrystallographer-style output an ideal pivot representation, enabling unambiguous, bidirectional translation between text and crystal. Editing the pivot therefore directly controls the crystal structure, addressing both challenges described above.
As illustrated in Fig.~\ref{fig:teaser}(a), our workflow proceeds as follows. We first map the user's imprecise natural-language prompt to a precise pivot description using an LLM. We then employ a GNN-based diffusion model~\citep{Park2025Chemeleon} trained on paired (pivot, crystal) data to generate candidate structures. Through the pivot, we can interpret ambiguous or high-level requests, overcoming the limitation of restricted instruction formats (Issue (a)).
Moreover, the pivot enables true bidirectional refinement. After generating a candidate structure, we convert the crystal back into its pivot description and update it based on user feedback. The refined pivot is then decoded again into a new crystal structure. This closed-loop pipeline resolves the second limitation—one-directional text-only generation—and makes iterative, conversation-style crystal design possible.

\begin{figure}[t]
\centering
\includegraphics[width=1.0\textwidth]{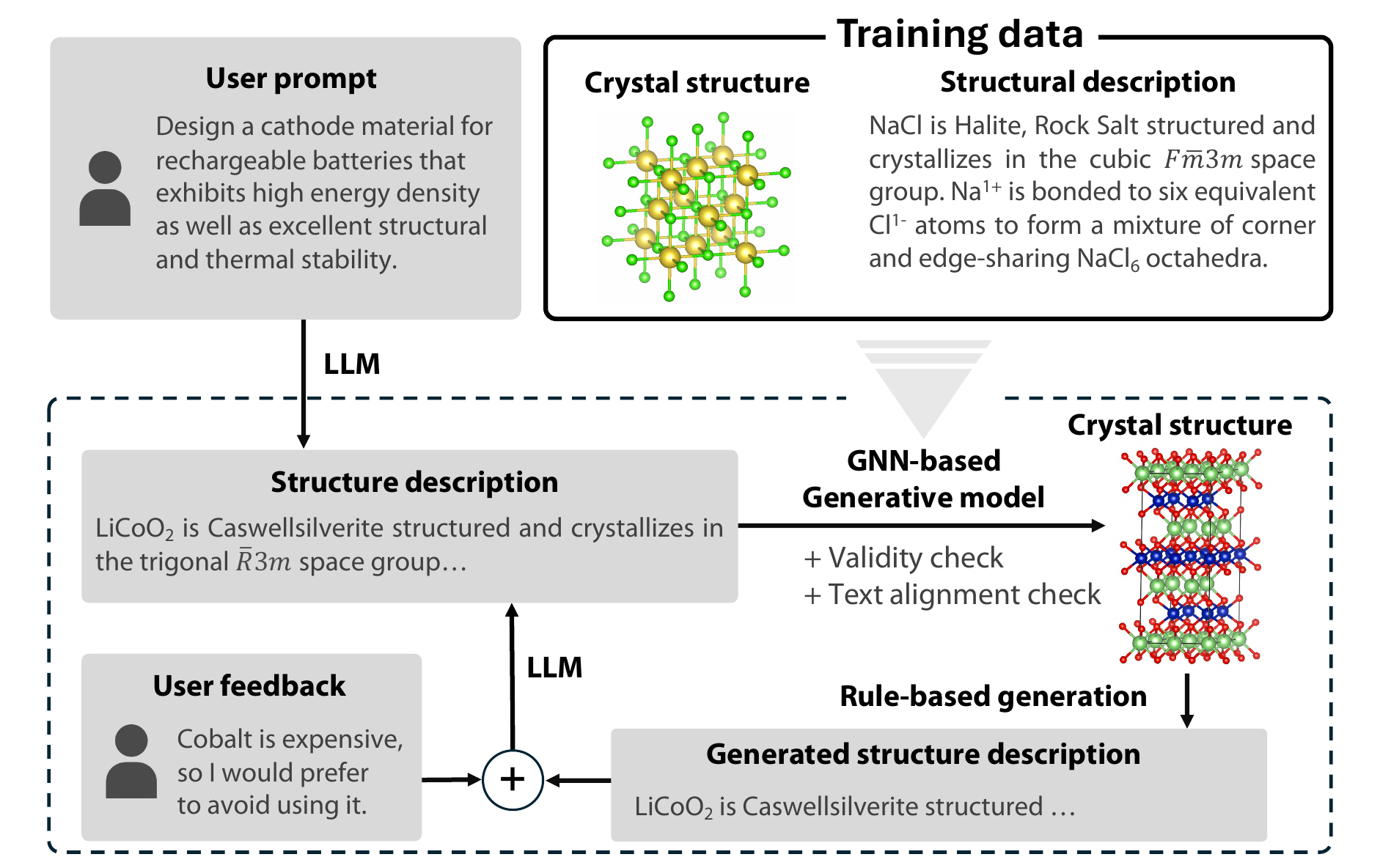}
\caption{
\textbf{Overview of \methodname, feedback-enabled framework for text-guided crystal structure generation.} 
The LLM interprets the prompt provided by the user into a pivot structure description, and the GNN-based generative model generates a crystal structure according to this description. Based on the previously generated crystal structure and the user’s feedback, the LLM then creates a description of the structure for the next generation. This approach enables the framework to leverage the LLM’s knowledge of materials science and the GNN-based model’s geometric reasoning capability.
}
\label{fig:iterative_framework}
\end{figure}

The full workflow is illustrated in Fig.~\ref{fig:iterative_framework}. To maximize generative quality, we adopt a Best-of-$N$ sampling strategy~\citep{karl2021training, nakano2021webgpt}: for each pivot description, the model generates $N$ candidate structures, verifies their physical plausibility (e.g., stability indicators, valid compositions), and selects the candidate most aligned with the input description.
After a structure is produced, it is presented to the user for feedback (\eg, comments like ``too distorted,'' ``replace zirconia with titanium'') or quantitative metrics (\eg, density, conductivity). The LLM receives this feedback alongside the pivot representation of the previously generated crystal and refines the pivot accordingly. A new crystal is then generated from the updated pivot, and the whole process repeats until the user is satisfied.

Through this iterative loop, the system progressively aligns the generated structures with the user's evolving desiderata. In this way, our framework mirrors—and directly augments—the traditional materials discovery workflow, which has long relied on repeated feedback and trial-and-error to refine candidate structures.
We provide details of the algorithm and its explanation in Sec.~\ref{s:methods_detail}.

\subsection{Quantitative analysis} \label{s:quantitative_analysis}
\begin{figure}[t]
    \centering
    \includegraphics[width=1.0\textwidth]{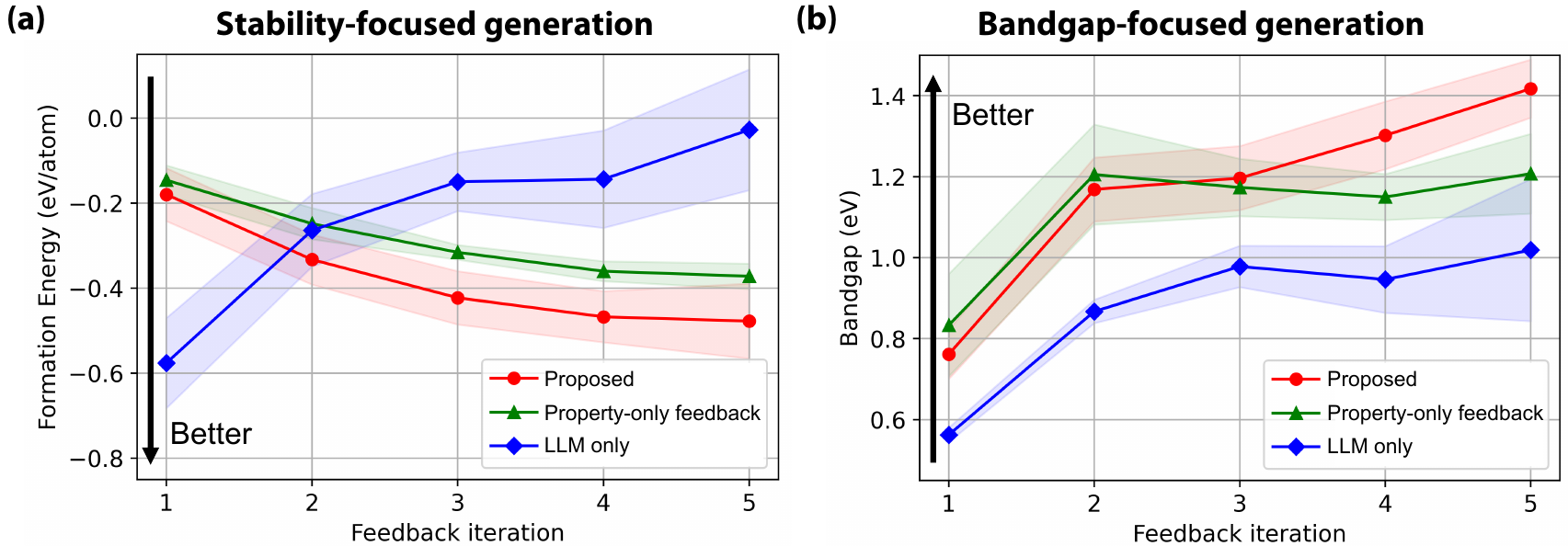}
    \caption{
    \textbf{Average formation energy and bandgap with standard deviation at each feedback iteration for (a) stability-focused generation and (b) bandgap-focused generation.} For comparison, we also present the results from property-only feedback and those obtained when the LLM directly generates complete crystal structure information. The proposed method progressively improves the targeted properties across multiple iterations and outperforms the other approaches. These results show that it makes effective use of feedback and highlight the combined strengths of the LLM and the GNN crystal generator. Data points are connected by lines for visual clarity.
    }
    \label{fig:improvement}
\end{figure}
To quantitatively evaluate the effectiveness of the feedback mechanism, we tested \methodname~in two cases: (a) generation focusing on structural stability and (b) generation focusing on large bandgap. 
Two types of prompts were used for generation: 
\begin{enumerate}
\renewcommand{\labelenumi}{(\alph{enumi})}
    \item \textbf{Stability-focused generation:} Generate a highly stable material with low formation energy that has not yet been reported. 
    \item \textbf{Bandgap-focused generation:} Generate a large bandgap material that has not yet been reported and is stable (i.e., has a low formation energy).
\end{enumerate}
For each prompt type, we generated crystal structures and conducted five rounds of feedback. We then analyzed the feedback trajectory, confirming that the formation energy progressively decreased in the stability-focused case and that the bandgap increased in the bandgap-focused case.

To enable rapid feedback cycles required for statistically reliable performance evaluation, we calculated material properties using an ML model and automatically generated feedback with an LLM. Specifically, we employed CrystalFramer~\citep{ito2025crystalframer}, a crystal property predictor trained on the MEGNet dataset~\citep{Chen2019megnet}, instead of performing computationally expensive DFT calculations~\citep{Hohenberg1964electrongas, kohn1965selfconsistent}. The pretrained weights provided by the authors were used for both formation energy and bandgap prediction.
The feedback processes were conducted using Alibaba's Qwen3-Next-80B-A3B-Thinking model~\citep{Yang2025qwen3}, and the prompt is provided in Appendix~\ref{a:prompt_feedback}. The LLM was provided with the generated structure’s Robocrystallographer description and the property prediction results from the ML model, and it was then prompted to give feedback based on them. We also adopted the same Qwen3 model for the generating process. We used the LLMs without any fine-tuning, and we selected Qwen3 since it runs locally, allows the environment to stay consistent for reproducible results, and is released under the Apache 2.0 license, which makes it easy to use. Please note that while the same model was used, the feedback and generation sessions were conducted separately, without shared conversation logs, to replicate human feedback conditions.

Previous methods generate crystal structures in a single step without incorporating feedback. Accordingly, the case with one feedback iteration corresponds to the baseline. We further compare our method with two additional baseline settings. \textbf{Property-only feedback}: We replace the LLM feedback with only the property values predicted by CrystalFramer, to assess whether the model can leverage the linguistic feedback. \textbf{LLM only}: we ask the LLM to directly generate complete crystal structures in the Crystallographic Information File (CIF) format, which is a domain-specific text representation of the full three-dimensional atomic configuration. This baseline corresponds to a simple approach where the entire generation process, including atomic coordinates, is carried out only in the language space without using the GNN-based generator, which is better suited to capturing complex three-dimensional atomic arrangements.

Figure~\ref{fig:improvement} shows the average formation energy and bandgap at each iteration for stability-focused generation (Panel (a)) and bandgap-focused generation (Panel (b)), where the values represent the mean and standard deviation obtained from five sets of 200-generation runs. In both cases, the proposed method outperformed the property-only feedback case, indicating that it effectively leveraged feedback to generate improved structures in subsequent iterations. The improvement observed even with property-only feedback case may result from the LLM producing new structures based on its internal knowledge, a behavior also noted in studies on self-refinement in language models~\citep{madaan2023selfrefine}.

When crystal structures were generated solely by the LLM (with CIF format information produced directly by the model), the stability-focused generation showed a gradual increase in formation energy over iterations, indicating decreasing stability, and the bandgap-focused generation performed less effectively than the proposed method. These differences highlight the complementary strengths of the LLM, which expresses structural descriptions linguistically, and the GNN, which captures the geometric characteristics of crystal structures. 

We also verified the stability of the bandgap-focused generation using the formation energies predicted by CrystalFramer~\citep{ito2025crystalframer}. Structures with predicted formation energies below 0~eV/atom were considered stable, and we calculated the proportion of stable structures among 1,000 generated samples. The proposed method achieved a stability rate of 77.2\%, and the property-only feedback setting reached 77.8\%. In contrast, the LLM that generated CIF files directly achieved 50.2\%. These results indicate that the proposed method satisfies both the requirement for an increased bandgap and the requirement for structural stability with high accuracy.

Furthermore, we assessed the generative capability of the model. As the structural and compositional filtering ensured the physical validity of the generated structures, we checked two additional key metrics, \textit{uniqueness} and \textit{novelty}. \textit{Uniqueness} represents the ratio of distinct structures obtained after removing duplicates using the \texttt{StructureMatcher} module in \texttt{pymatgen}~\citep{ONG2013pymatgen}. \textit{Novelty} denotes the proportion of generated structures for which no similar structures exist among the 210,579 entries of the Materials Project database as of October 2025~\citep{Horton2025MP}, as determined by \texttt{StructureMatcher}. 

Table~\ref{tab:uniqueness_novelty} summarizes the uniqueness and novelty observed in each case. Regarding uniqueness, in the methods combined with GNN, almost all of the 1,000 generated structures were distinct, while in the case where the LLM directly generated complete crystal structure information, the rate was in the 80\% range. These differences suggest that when the LLM is allowed to output structural descriptions in natural language rather than directly producing specialized formats such as CIF, it can more fully leverage its generative capability. As for novelty, the rate was 100\% in all cases. Since the prompts explicitly instructed the model to create new structures, it seems that the LLM followed those instructions faithfully. Our evaluation was conducted using only the Materials Project dataset, and we acknowledge that the LLM’s internal knowledge extends beyond this source; therefore, the assessment is not entirely exhaustive. Nevertheless, we confirmed that the generated structures were not ones that are widely known.

\begin{table}[t]
\centering
\caption{\textbf{Uniqueness and novelty of generated crystal structures.}}
\begin{tabular}{llcc}
\toprule
\textbf{Generation prompt} & \textbf{Method} & \textbf{Uniqueness (\%)} & \textbf{Novelty (\%)} \\
\midrule
\multirow{3}{*}{Stability-focused}  & Proposed             & 98.9 & 100 \\
                                    & property-only feedback & 98.4 & 100 \\
                                    & LLM only & 87.2 & 100 \\
\cmidrule(lr){1-4}
\multirow{3}{*}{Bandgap-focused} & Proposed             & 99.8 & 100 \\
                                    & property-only feedback     & 99.7 & 100 \\
                                    & LLM only & 81.8 & 100 \\
\bottomrule
\end{tabular}
\label{tab:uniqueness_novelty}
\end{table}

\subsection{Qualitative analysis}
\label{s:qualitative_analysis}
\begin{figure}[t]
\centering
\includegraphics[width=0.95\textwidth]{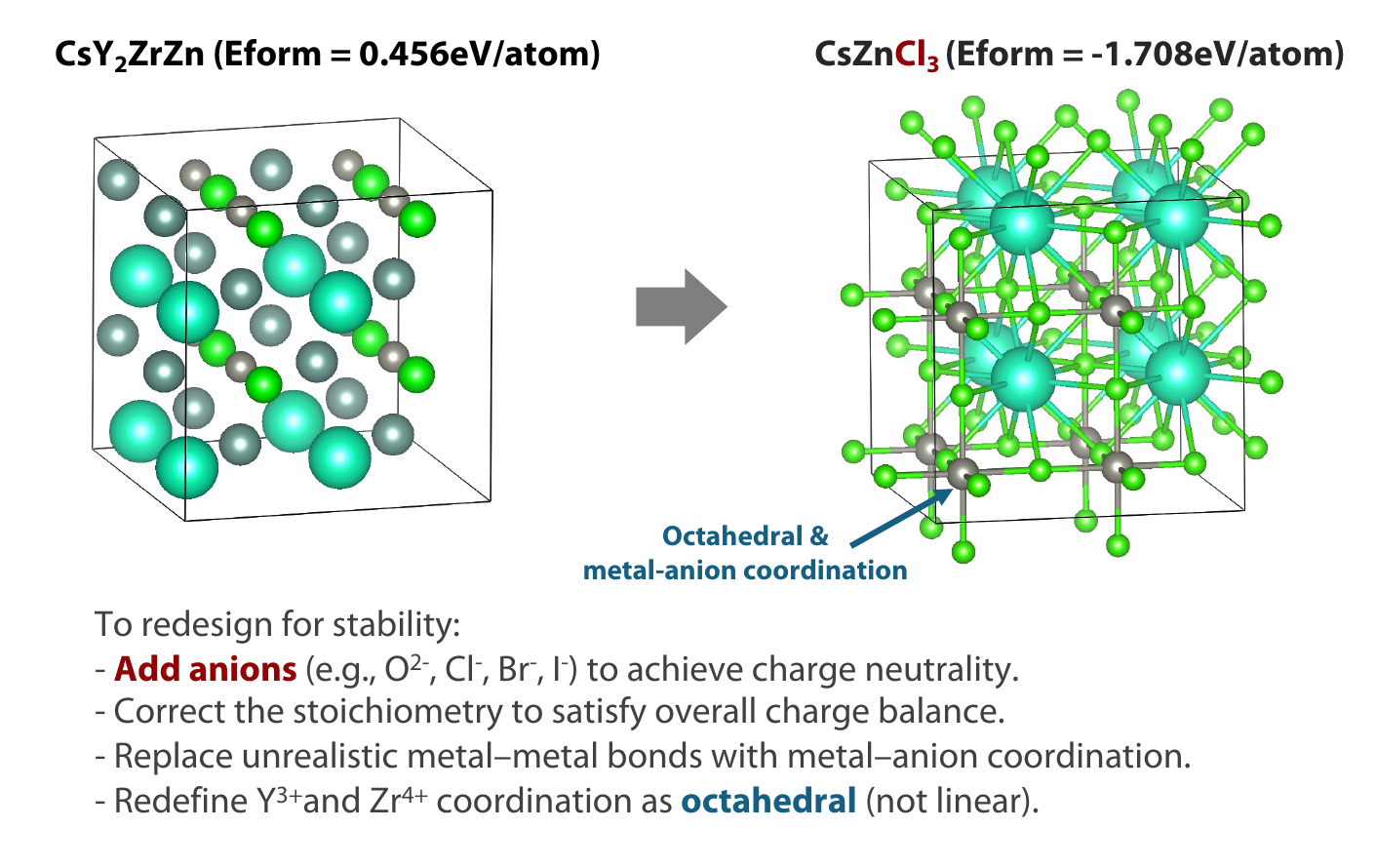}
\caption{
\textbf{An Example of structure refinement through the feedback process.}
The model successfully incorporated linguistic feedback into both compositional and structural refinements. Crystal structures are visualized by Vesta~\citep{Momma2011vesta}. 
}
\label{fig:example_refinement}
\end{figure}

In addition to the quantitative evaluation presented in Sec.~\ref{s:quantitative_analysis}, we qualitatively examined the feedback obtained at each iteration and analyzed how the structure evolved accordingly. Figure~\ref{fig:example_refinement} presents an example of the stability-focused generation task described in Sec.~\ref{s:quantitative_analysis}, showing the structures before and after feedback together with a summary of the corresponding comments. 

In the example shown in Fig.~\ref{fig:example_refinement}, the initially generated structure was pointed out to lack anions, and it was suggested to replace the unrealistic metal–metal bonds with metal–anion bonds. In addition, it was recommended that metallic atoms such as ytterbium and zirconium adopt an octahedral coordination rather than a linear one. Reflecting these suggestions, the refined structure forms a halide perovskite structure, where the previously identified issues have been resolved, and the predicted formation energy improved from 0.456~eV/atom to –1.708~eV/atom. From this result, it is evident that the received feedback was correctly reflected in the generated structures. In particular, it is remarkable that not only elemental substitutions but also structural modifications were accurately performed. 

We also demonstrated in Fig.~\ref{fig:teaser}(b) that crystal structures can be designed not only through organized feedback from an LLM but also through human colloquial feedback. Using the insulator discovery task as an example, we illustrated how the crystal structure evolved over three rounds of feedback. In the first round, we simply provided predicted property values obtained from the crystal structure encoder. The model then modified the structure while preserving the composition, successfully widening the bandgap. In the second round, we instructed the model to replace strontium with a more readily available element. Although the bandgap slightly decreased, strontium was replaced with calcium, which is abundant on Earth. In the third round, we asked the model to increase the distance between structural units to disrupt conduction pathways. The model responded by editing the structure while maintaining the composition, resulting in enlarged inter unit distances. Overall, these results demonstrate that our framework supports an iterative and conversational design process, in which initially vague user desiderata are gradually refined into more concrete design constraints, including restrictions on elemental composition, rather than being treated as a fixed property optimization problem.

\subsection{Crystal editing based on textual instructions}
\begin{figure}[t]
\centering
\includegraphics[width=0.85\textwidth]{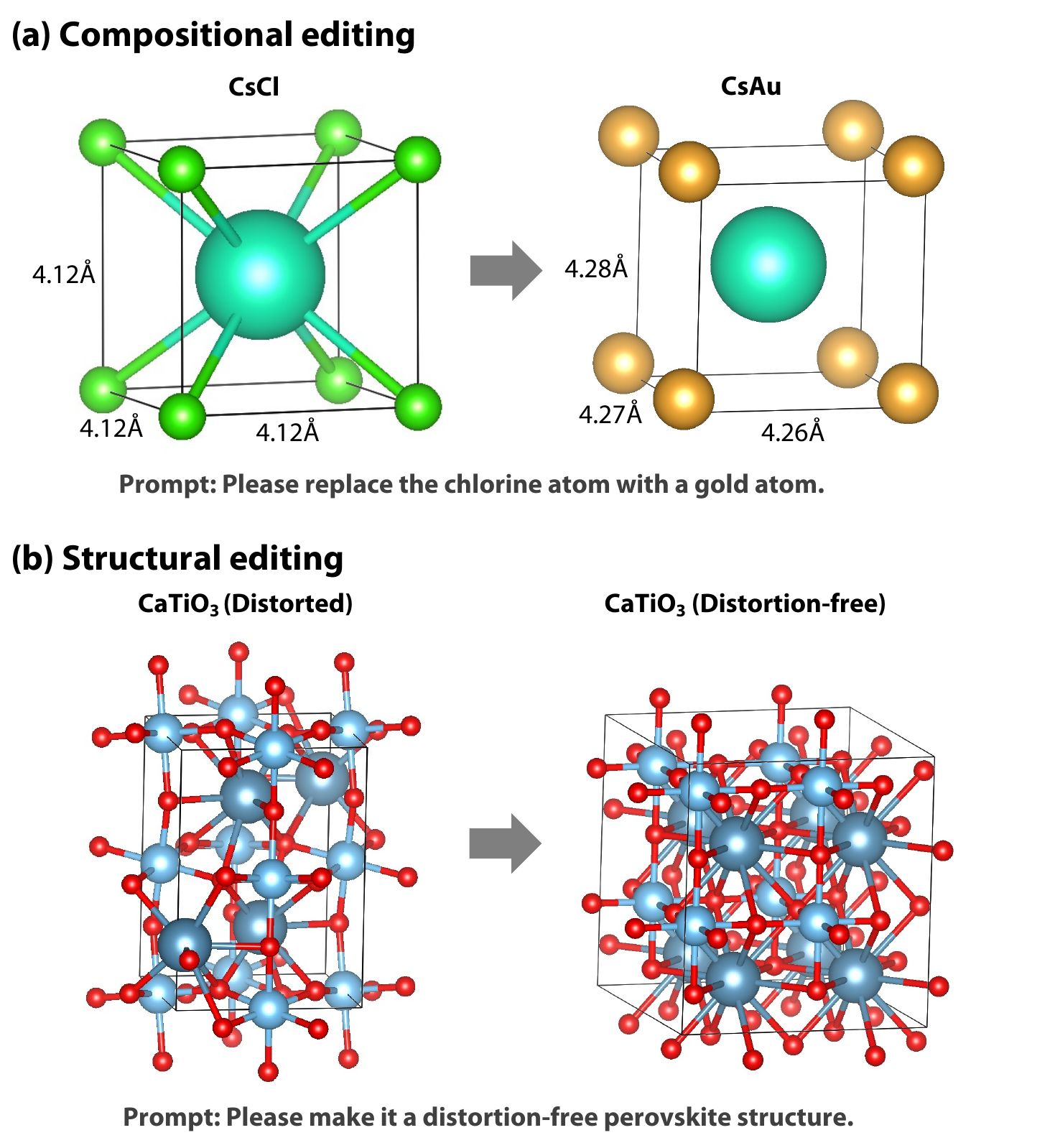}
\caption{
\textbf{Crystal editing demonstrations: (a) Elemental substitution in CsCl and (b) Structural editing in a perovskite structure.} 
We confirmed that the model successfully edited the crystal structures in accordance with the text instructions and adjusted structural parameters such as lattice constants to achieve more stable configurations. Crystal structures are visualized by Vesta~\citep{Momma2011vesta}. 
}
\label{fig:crystal2crystal}
\end{figure}
While Fig.~\ref{fig:iterative_framework} and Algorithm~\ref{alg:iterative_generation} show a process that begins with a text prompt to generate a crystal structure from scratch, \methodname~also enables crystal editing, creating new structures from existing ones under textual guidance. This approach parallels image-to-image generation, where an image is modified according to text prompts~\citep{Guillaume2022DiffEdit, chang2023Muse, kawar2023imagic}. 
In this setting, the generation process is initialized with an existing crystal structure and a feedback prompt, and the procedure then starts from the upper-right part of the feedback loop in Fig.~\ref{fig:iterative_framework}~(i.e., line 10 of Algorithm~\ref{alg:iterative_generation}).

We conducted the crystal editing experiment in two cases: (a) a relatively simple element-substitution case and (b) a structural-modification case. In the former, we instructed the model to replace the chlorine atoms in the CsCl structure with gold atoms. In the latter, we performed an editing task to transform the distorted perovskite structure of CaTiO$_3$ into an undistorted perovskite structure. 

In the compositional editing shown in Fig.~\ref{fig:crystal2crystal}(a), we confirmed that the chlorine atoms are correctly substituted with gold atoms. Interestingly, this substitution results in a change in the lattice constant from 4.12~\AA\ to approximately 4.27~\AA, with the generated CsAu exhibiting a larger value than CsCl. This trend is consistent with previous experimental reports, which give lattice constants of 4.12~\AA\ for CsCl and 4.26~\AA\ for CsAu~\citep{Ralph1921CsCl, Tinelli1978CsAu}, and suggests that lattice expansion accompanies structural stabilization.

As for the structural editing results shown in Fig.~\ref{fig:crystal2crystal}(b), the distorted perovskite structure of CaTiO$_3$ was successfully transformed into a distortion-free, regular perovskite structure, which indicates that our model is capable of capturing and correcting geometric structural information.

From these results, we verified that \methodname~can intuitively modify crystal structures based on textual instructions given for the original structure. It was also confirmed that the model not only follows the instructions but also adjusts structures such as lattice constants to achieve a more stable crystal configuration.

\section{Discussion}
In this study, we demonstrated that introducing a linguistic description of structure as an intermediate representation makes it possible to connect text and crystal structures in a feedback-capable manner. We adopted the notation of Robocrystallographer as the pivot representation, as it provides the most intuitive linguistic expression~\citep{Ganose2019Robocrystallographer}. However, other approaches such as SLICES~\citep{Xiao2023SLICES} have also been developed to describe crystal structures linguistically~\citep{jia2025benchmarking}. A comparative investigation of these representations will be left for future work.

The crystal structure generation module of our proposed framework is based on Chemeleon~\citep{Park2025Chemeleon} and does not impose any symmetry constraints. 
As a result, most of the generated structures are classified in the lowest space group \textit{P1}, although they often roughly satisfy higher crystallographic symmetries. The remaining deviations from exact symmetry reflect the weak sensitivity of the loss function to strict enforcement of these symmetries.
For instance, in the generated structure shown in Fig.~\ref{fig:crystal2crystal}(a), the lattice parameters are approximately equal in length but not identical.
Recently, considerable progress has been made in crystal generation methods that explicitly incorporate symmetry~\citep{jiao2024diffcsp, Zhu2024wycryst, levy2025symmcd, kelvinius25wyckoffdiff, kazeev2025wyckofftransformer}. Extending these advances toward symmetry-aware text-to-crystal structure generation remains an area for future work. Another promising approach is to incorporate postprocessing methods, such as DFT relaxation and symmetry detection~\cite{togo2024spglib} with large tolerance to correct subtle discrepancies. Additionally, symmetry itself has a hierarchical nature, and just as symmetry decreases during phase transitions, natural structural variations in crystals should also follow this hierarchy~\citep{Englert2013symmetryrelation}. Integrating this hierarchical aspect of symmetry into the iterative generation process (\eg, allowing the next generation step to adopt symmetry one level lower or higher than that of the original structure) will be an important direction for future research.

In the quantitative evaluation in Sec.~\ref{s:quantitative_analysis}, although the feedback came from an LLM rather than human experts, the predicted physical properties shifted toward more favorable values. Similar performance improvements through LLM feedback have been reported in other domains~\citep{madaan2023selfrefine}, and it is intriguing that the same effect was also observed in the highly complex task of crystal generation. This ability to autonomously explore high-quality crystal structures indicates that the proposed approach can serve as a component within an agent-based workflow for materials discovery. When combined with the rapidly growing field of autonomous experimentation, it becomes a powerful engine for progress in crystal design and marks an important step toward a self-driving laboratory that can independently iterate design, synthesis, and evaluation \citep{Burger2020mobilerobotic, Tom2024selfdriving}.

In this study, we focused on quantifiable properties such as formation energy and bandgap, which can be evaluated through DFT calculations or their surrogate machine learning models. Meanwhile, the performance required in practical materials development is often not something that can be directly and quantitatively assessed; rather, it tends to involve a combination of diverse and sometimes qualitative requirements. Demonstrating our method in an actual materials development context would demand advanced expertise and practical knowledge specific to the field, and thus was not performed in this work. The potential of this approach to extend to complex materials development remains an interesting open question.

\section{Methods} \label{s:methods}
\subsection{Details of \methodname}
\label{s:methods_detail}
Algorithm~\ref{alg:iterative_generation} summarizes the proposed framework described in Sec.~\ref{s:pivot_representation}. The initially provided target property description $p_1$ is converted into a Robocrystallographer-format structural description $d_1$ by using an LLM, and a pure noise state is prepared for crystal generation via a GNN-based diffusion model. Then the loop begins, consisting of two main stages. The first stage (lines 4--7) generates a crystal structure from the structural description, while the second stage (lines 8--13) receives feedback on the generated crystal structure and, based on the result, produces the next structural description and the noise state to be denoised during generation. Each stage is described in detail in the following Sec.~\ref{s:generative_model} and Sec.~\ref{s:feedback_strategy}.

\subsubsection{Crystal structure generation from structural descriptions}
\label{s:generative_model}

\begin{algorithm}[t]
\caption{\methodname}
\label{alg:iterative_generation}
\begin{algorithmic}[1]
\Require
  \Statex \textbf{Input:} 
  \Statex \hspace{1.5em} Target property text $p_1$ (\eg, ``high electrical conductivity'')
  \Statex \hspace{1.5em} Structural description-conditioned crystal diffusion model \ensuremath{\mathcal{G}}
  \Statex \hspace{1.5em} Number of feedback iterations $K$
  \Statex \hspace{1.5em} Number of generations per iteration $N$ (default: 10)
  \Statex \hspace{1.5em} Denoising strength $\alpha \in(0,1]$ (default: 0.1)
\Ensure Optimized crystal structure $c^*$

\Statex
\State $d_1 \gets \texttt{LLM\_interpret}(p_1)$ \Comment{Details in Appendix~\ref{a:prompt_input}}
\State Initialize $\{\mathcal{Z}_i\}_{i=1}^N$ with a pure noise state for the diffusion model
\For{$k = 1$ to $K$}
    \State \textbf{Generate structure $c_k$ from structural description:} \Comment{Details in Sec.~\ref{s:generative_model}}
    \State \hspace{1.5em} Sample $N$ candidate structures $\{c_k^{(i)}\}_{i=1}^N$ using the diffusion model $\mathcal{G}$
    \[
        c_k^{(i)} = \mathcal{G}(\mathcal{Z}_i, \text{condition}=d_k)
    \]
    \State \hspace{1.5em} Evaluate structural and compositional validity, and filter out invalid samples:
    \[
        \{c_k^{(i)}\} \gets \{c_k^{(i)} \mid \texttt{isValid}(c_k^{(i)})=\texttt{True}\}
    \]
    \State \hspace{1.5em} Compute text–structure alignment scores and pick best:
    \[
        c_k \gets \arg\max_i \texttt{AlignmentScore}(c_k^{(i)}, d_k)
    \]

    \State \textbf{Feedback and refinement:} \Comment{Details in Sec.~\ref{s:feedback_strategy}}
    \State \hspace{1.5em} Provide feedback based on the generated structure $p_k^{\text{fb}} \gets \texttt{Feedback}(c_k)$
    \State \hspace{1.5em} Convert $c_k$ to structural description $d^{\text{gen}}_k \gets \texttt{Robocrystallographer}(c_k)$
    \State \hspace{1.5em} Update structural description with feedback $p^\text{fb}_k$:
    \[
        d_{k+1} \gets \texttt{LLM\_refine}(p^\text{fb}_k, d^\text{gen}_k)
    \]
    \State \hspace{1.5em} Initialize next step from partially noised state:
    \[
        \{\mathcal{Z}_i\}_{i=1}^N \gets \texttt{Add\_noise}(c_k, \text{strength}=\alpha)
    \]
    \State \hspace{1.5em} Denoise from $\{\mathcal{Z}_i\}_{i=1}^N$ in the next iteration
\EndFor
\State \Return $c^* \gets c_K$
\end{algorithmic}
\end{algorithm}

We employed Chemeleon~\citep{Park2025Chemeleon} to generate crystal structures from Robocrystallographer format descriptions~\citep{Ganose2019Robocrystallographer}. Details of the model are provided in Sec.~\ref{s:model_detail}. To encourage the model to generate crystals that are physically plausible and more faithful to the given text, we generated $N=10$ candidate samples (\ie, each starting from a different pure noise state). Among them, only the samples satisfying both structural and compositional validity were retained, following the validity criteria commonly used in previous studies~\citep{xie2021cdvae, jiao2024diffcsp, levy2025symmcd}. Specifically, structural validity was determined by ensuring that no pair of atoms was closer than 0.5 \AA, while compositional validity was checked by confirming overall charge neutrality using the SMACT library~\citep{Davies2019smact}. 
Finally, the structure with the highest alignment score—analogous to the CLIP score~\citep{hessel2021clipscore}, which indicates the degree of semantic consistency between the generated structure and the text description—was selected as the final output.
This alignment score was computed by encoding the crystal structures and the text with encoders trained during the contrastive learning stage of Chemeleon (see Sec.~\ref{s:model_detail}), and then measuring the cosine similarity between their embeddings. Note that if none of the structures passes the validity check, the iteration is reset and repeated.

\subsubsection{Crystal structure refinement}
\label{s:feedback_strategy}
To design the next crystal structure, feedback $p_k^{\text{fb}} = \texttt{Feedback}(c_k)$ is provided based on the generated crystal structure $c_k$. Although only $c_k$ appears as an argument, it should be noted that the feedback is not derived solely from $c_k$ itself, but from various results (\eg, simulation and experimental results).
We then obtain the Robocrystallographer representation of the previously generated crystal structure and supply it to the LLM along with feedback, as illustrated in the prompt shown in Appendix~\ref{a:prompt_refine}. Rather than initiating the next crystal structure generation from a pure noise state, the diffusion process starts from a partially noised version of the original structure, ensuring that the refinement stays guided by the initial configuration. We control the level of noise through the denoising strength parameter $\alpha \in (0,1]$, a technique commonly used in image-to-image tasks~\citep{Guillaume2022DiffEdit}.

In this approach, crystal structure generation from feedback begins not from a pure noise state but from a partially noisy state of the previously generated crystal structure corresponding to the diffusion time step $T \times \alpha$, where $T$ denotes the fully noisy state and $\alpha$ controls the diffusion strength. This partial noising process enables the model to refine or adjust the output while preserving its overall character.

We examined the effect of the strength parameter $\alpha$ in Appendix~\ref{a:strength_effect}.

\subsection{Crystal structure generation model} \label{s:model_detail}
We strictly followed Chemeleon for the model that generates crystal structures from linguistic structural descriptions. After briefly describing the model architecture, we provide a detailed explanation of the dataset and training details. For a more comprehensive description of the architecture, please refer to the original paper~\citep{Park2025Chemeleon}.
\subsubsection{Model architecture}
Chemeleon is a text-guided generative model that generates crystal structures through a denoising diffusion process conditioned on text embeddings from a pretrained text encoder.
During training, Gaussian noise is added to crystal structures, and the model is trained to predict the added noise. 
During inference, the model starts from a pure noise state and progressively denoises it to generate complete crystal structures. For the lattice constants and atomic positions, it follows the framework of Denoising Diffusion Probabilistic Models (DDPM)~\citep{ho2020diffusion}, while for the atomic species, it adopts the Discrete Denoising Diffusion Probabilistic Models (D3PM) framework~\citep{austin2021d3pm}.

Chemeleon comprises two key elements. The first component is Crystal CLIP, a cross-modal contrastive learning module for pretraining the text encoder MatTPUSciBERT~\citep{Gupta2022matscibert} by aligning its embeddings with the corresponding crystal structure embeddings produced by the GNN. By bringing positive text–crystal pairs closer together and pushing negative pairs farther apart, Crystal CLIP learns a shared latent space where textual representations reflect structural geometric information.

The second element is a classifier-free guided denoising diffusion model~\citep{ho2022cfg} that predicts the noise added to each variable of the crystal structure (\ie, lattice matrices, atomic coordinates, and atom types), conditioned on the text embeddings produced by the text encoder of Crystal CLIP. The denoising network builds upon the DiffCSP framework~\citep{jiao2023diffcsp}, which was originally developed for crystal structure prediction tasks.

By aligning linguistic embeddings with the geometric information of crystal structures through CLIP and training the denoising model conditioned on these embeddings, the model can generate crystal structures that follow textual instructions.

\subsubsection{Dataset and training details}
We used the MEGNet dataset~\citep{Chen2019megnet}, which is a snapshot of the Materials Project database~\citep{Horton2025MP}. Following the official split, the dataset was divided into 60,000, 5,000, and 4,239 samples for training, validation, and testing, respectively. After generating textual descriptions using Robocrystallographer, we trained the model on this dataset. 

The training of Chemeleon consists of two stages: (1) contrastive pretraining of the Crystal CLIP module, and (2) text-conditioned diffusion model training for crystal generation.

In the contrastive learning stage, the text encoder and the GNN-based crystal encoder are trained together so that their embeddings align within a shared latent space, which helps the text embedding capture geometric information. The text embeddings are obtained from the \texttt{[CLS]} token of the text encoder output, and the crystal embeddings are produced by averaging the node features from the GNN-based crystal structure encoder. The training objective combines text-to-graph and graph-to-text cross-entropy losses with a symmetric contrastive formulation. A batch size of 128 is used with the Adam optimizer~\citep{kingma14adam}, where the learning rates for the text and graph encoders are set to $1\times 10^{-5}$ and $1\times 10^{-4}$, respectively. Training proceeds for up to 1,000 epochs, employing early stopping if the validation loss does not improve for 300 epochs. A learning rate scheduler with \texttt{ReduceLROnPlateau} (patience = 200 epochs) is applied for stability.

During diffusion model training, the text encoder is kept frozen, and the pretrained Crystal CLIP embeddings are used as conditional inputs. The denoising network is optimized using the Adam optimizer with a learning rate of $1\times 10^{-3}$, maintaining the same batch size and scheduling settings as in the contrastive learning stage. The loss function consists of three components: atom species, lattice, and coordinate denoising losses. Both training stages were performed on four NVIDIA H200 (141 GB) GPUs and took 30 hours for contrastive learning and 20 hours for diffusion model training.

\section*{Competing interests}
All authors declare no financial or non-financial competing interests. 

\section*{Author contributions}
\textbf{Y.I.} initiated the study, designed and implemented the method, conducted all numerical and experimental analyses, and drafted the initial manuscript.
\textbf{Y.S.} supervised the study, provided advice on the method and experimental design, and reviewed the draft manuscript.
\textbf{T.M.} provided advice on the method and experimental design and reviewed the draft manuscript.
\textbf{M.A.} revised the manuscript and provided overall project supervision, including resources, funding, and institutional support.

\bibliography{main}
\bibliographystyle{naturemag-doi-eprint.bst}

\newpage

\appendix
\thispagestyle{plain}
\renewcommand{\thefigure}{A.\arabic{figure}}
\renewcommand{\theequation}{A.\arabic{equation}}
\renewcommand{\thetable}{A.\arabic{table}}
\renewcommand{\thealgorithm}{A.\arabic{algorithm}}
\setcounter{equation}{0}
\setcounter{figure}{0}
\setcounter{table}{0}
\setcounter{algorithm}{0}
\renewcommand{\theHfigure}{A\arabic{figure}}
\renewcommand{\theHtable}{A\arabic{table}}
\renewcommand{\theHequation}{A\arabic{equation}}
\renewcommand{\theHalgorithm}{A\arabic{algorithm}}

\begin{center}
    \huge{\textbf{Appendix}}
\end{center}

\section{Prompt for feedback} \label{a:prompt_feedback}
We used the following prompt for feedback. Although this step is intended to be performed by an expert, we conducted it using an LLM for quantitative evaluation.
\begin{tcolorbox}[colback=gray!5,colframe=black!50,title=Prompt for feedback]
You are a scientist specializing in materials science with expertise in designing crystal structures.  \\
I want to create the following material: ``\textit{prompt} $p_1$''. \\ 
I will provide the crystal structure and its property, so please give me advice on how to update the crystal structure. 
\end{tcolorbox}

\section{Prompt for initial structural description from property} \label{a:prompt_input}
The following prompt was adopted for $\texttt{LLM\_interpret}$, which generates structural descriptions of crystal structures from user input. Since the crystal structure generation model requires a predefined number of atoms, we instructed the LLM to output this information using the $\texttt{--n\_atoms}$ tag.

\begin{tcolorbox}[colback=gray!5,colframe=black!50,title=Prompt for LLM\_interpret]
You are a scientist specializing in materials science with expertise in designing crystal structures.  

Describe the structural features typically associated with materials that exhibit the specified physical property. Provide detailed information on crystal symmetry, lattice type, bonding characteristics, coordination environments, and common structural motifs. Present the output in **Markdown format**. \\

**Instructions:**  

- Follow the example format exactly.  

- Do not use general chemical formulas (e.g., ABO$_3$).  

- Responses that do not follow the format will be considered incorrect.  

- Carefully consider what type of crystal structure is necessary to realize the given property.  

- At the end of the description, return the number of atoms in the unit cell using the tag format -~-n\_atoms=integer.  

- Assume that I will attempt to synthesize the proposed structure and evaluate its physical properties. \\

**Example**  

BaTiO$_3$ adopts a cubic perovskite structure and crystallizes in the cubic space group Pm-3m. Ba$^{2+}$ is coordinated by twelve equivalent O$^{2-}$ atoms to form BaO$_{12}$ cuboctahedra, which share corners with twelve equivalent BaO$_{12}$ cuboctahedra, faces with six equivalent BaO$_{12}$ cuboctahedra, and faces with eight equivalent TiO$_6$ octahedra. All Ba--O bond lengths are 2.83~\AA. Ti$^{4+}$ is coordinated by six equivalent O$^{2-}$ atoms to form TiO$_6$ octahedra, which share corners with six equivalent TiO$_6$ octahedra and faces with eight equivalent BaO$_{12}$ cuboctahedra. The corner-sharing TiO$_6$ octahedra are not tilted. All Ti--O bond lengths are 2.00~\AA. Each O$^{2-}$ is bonded in a distorted linear coordination to four equivalent Ba$^{2+}$ and two equivalent Ti$^{4+}$ atoms.  

-~-n\_atoms=20
\end{tcolorbox}

\newpage
\section{Prompt for refining structural descriptions} \label{a:prompt_refine}
To refine the structural description, we used the following prompt for $\texttt{LLM\_refine}$. The \textit{italicized parts} should be replaced with the corresponding descriptions of the generated crystal structures.

\begin{tcolorbox}[colback=gray!5,colframe=black!50,title=Prompt for LLM\_refine]
The synthesis of the crystal structure you proposed yielded ``\textit{Formula}'', whose structure is as follows: ``\textit{Robocrystallographer's description} $d_k^\text{gen}$.''

-- 

``\textit{User Instruction} $p_k^\text{fb}.$''

--

Can you refine the crystal structure? Please respond full crystal structure description in accordance with the format specified in the instructions. Do not return any other things.

\end{tcolorbox}


\section{Effect of diffusion strength} \label{a:strength_effect}
We examined the effect of the diffusion strength parameter $\alpha$ introduced in Sec.~\ref{s:methods_detail} by measuring the \textit{success rate}, defined as the proportion of cases in which at least one valid crystal structure appeared in each of the five iterations. The validity of each structure was evaluated using the same method described in line 6 of Algorithm~\ref{alg:iterative_generation}. In each iteration, 30 crystal structures were generated. For each value of the strength parameter, we conducted 300 trials for stability-focused generation.

\begin{figure}[h]
\centering
\includegraphics[width=1.0\textwidth]{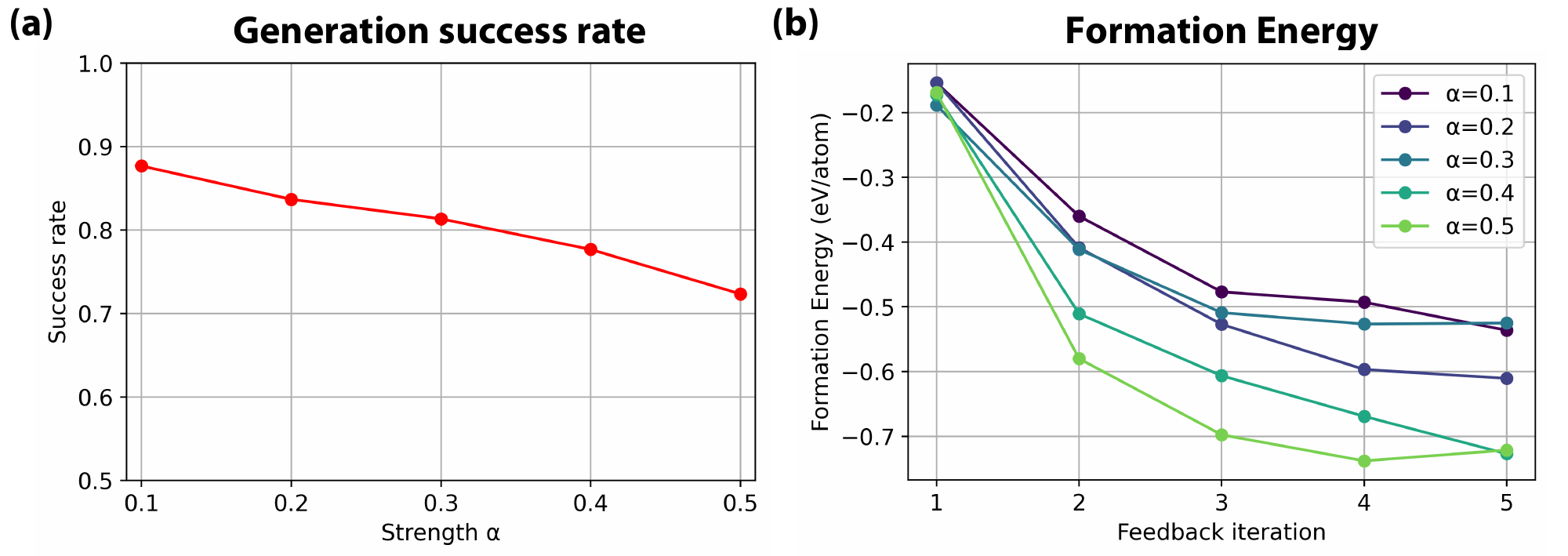}
\caption{
\textbf{Comparison of generation performance under different diffusion strengths.}
Panels (a) and (b) show the effects of varying diffusion strength $\alpha$ on the generation success rate, and mean formation energy, respectively.
(a) The generation success rate decreases as $\alpha$ increases, indicating that weaker influence from the previous step leads to greater structural variation and makes it more difficult to pass the validity check.
(b) Mean formation energy values improve with larger $\alpha$, suggesting that reduced dependence on previous structures promotes the discovery of more favorable configurations.
}
\label{fig:strength_comparison}
\end{figure}

Figure \ref{fig:strength_comparison}(a) shows the dependence of the success rate on the diffusion strength, and Panel (b) illustrates the evolution of the average formation energy through the iterations. 
Figure~\ref{fig:strength_comparison}(a) shows a clear tendency for the success rate to decrease as the diffusion strength $\alpha$ increases. This trend can be attributed to the reduced influence of structural information from the previous step at each iteration as $\alpha$ increases, which leads to greater structural variation and makes it more difficult to pass the validity check. In contrast, the mean values of the formation energy tended to improve with increasing $\alpha$, as shown in Fig.~\ref{fig:strength_comparison}(b). These results suggest that weakening the dependence on the previous structure allows the exploration of new favorable structures, while strengthening the dependence steers the process toward exploitation. Therefore, the value of strength $\alpha$ should be carefully tuned according to the goal, whether one aims to make only minor structural modifications or to discover entirely new structures.

\end{document}

%% file: math_commands.tex
\usepackage{amsmath,amsfonts,bm}

\def\eqref#1{equation~\ref{#1}}

\def\1{\bm{1}}

\DeclareMathAlphabet{\mathsfit}{\encodingdefault}{\sfdefault}{m}{sl}
\SetMathAlphabet{\mathsfit}{bold}{\encodingdefault}{\sfdefault}{bx}{n}

